\documentclass[sigconf, screen]{acmart}
\usepackage{booktabs}
\usepackage{multirow}
\AtBeginDocument{
  }
\copyrightyear{2026}
\acmYear{2026}
\setcopyright{cc}
\setcctype{by-nc-nd}
\acmConference[MM '26]{Proceedings of the 34th ACM International Conference on Multimedia}{November 10--14, 2026}{Rio de Janeiro, Brazil}
\acmBooktitle{Proceedings of the 34th ACM International Conference on Multimedia (MM '26), November 10--14, 2026, Rio de Janeiro, Brazil}
\acmDOI{10.1145/3767308.3836278}
\acmISBN{979-8-4007-2213-4/2026/11}

\newcommand{\mypara}[1]{\smallskip\noindent{\bf {#1}.}}
\usepackage{enumitem}
\usepackage{cleveref}
\hypersetup{unicode=true}
\usepackage{pifont}
\newcommand{\cmark}{\ding{51}}
\newcommand{\xmark}{\ding{55}}
\acmSubmissionID{7660}

\begin{document}

\title{TransMeme: A Multi-Agent Framework for Cross-Cultural Meme Transcreation}

\author{Jingyi Zheng}
\affiliation{%
  \institution{Hong Kong University of Science and Technology (Guangzhou)}
  \city{Guangzhou}
  \country{China}}
\affiliation{%
  \institution{Wuhan University}
  \city{Wuhan}
  \country{China}}
\email{jzheng029@connect.hkust-gz.edu.cn}

\author{Yule Liu}
\affiliation{%
  \institution{Hong Kong University of Science and Technology (Guangzhou)}
  \city{Guangzhou}
  \country{China}}
\email{yliu514@connect.hkust-gz.edu.cn}

\author{Zifan Peng}
\affiliation{%
  \institution{Hong Kong University of Science and Technology (Guangzhou)}
  \city{Guangzhou}
  \country{China}}
\email{zpengao@connect.hkust-gz.edu.cn}

\author{Tianyi Hu}
\affiliation{%
  \institution{Aarhus University}
  \city{Aarhus}
  \country{Denmark}}
\email{tenney.hu@cs.au.dk}

\author{Yuemeng Zhao}
\affiliation{%
  \institution{Hong Kong University of Science and Technology (Guangzhou)}
  \city{Guangzhou}
  \country{China}}
\email{yzhao472@connect.hkust-gz.edu.cn}

\author{Xinhu Zheng}
\affiliation{%
  \institution{Hong Kong University of Science and Technology (Guangzhou)}
  \city{Guangzhou}
  \country{China}}
\email{xinhuzheng@hkust-gz.edu.cn}

\author{Xinlei He}
\correspondingauthor
\affiliation{%
  \institution{Wuhan University}
  \city{Wuhan}
  \country{China}}
\email{xinlei.he@whu.edu.cn}

\renewcommand{\shortauthors}{Jingyi Zheng et al.}
%% No italics, no superscripts, not anonymous
%% Use footnote or author note to identify equal contribution, shared contribution, and/or contact author info


\begin{abstract}
Internet memes are a pervasive form of multimodal online communication; however, such communication often involves users from diverse linguistic and cultural backgrounds. Therefore, adapting memes across cultures and languages is a central challenge for enabling mutual understanding in online communication.
Unlike ordinary translation or standalone text rewriting, cross-cultural meme transcreation must jointly preserve communicative intent, adapt culture-dependent meaning for the target audience, and maintain coherence between text and image. 
In this work, we first provide an explicit task analysis of cross-cultural meme transcreation and identify three core challenges: \textit{Culture-specific knowledge understanding}, \textit{Intent and Tone Preservation}, and \textit{Multimodal Consistency}. 
Based on this analysis, we propose a multi-agent framework with specialized agents that are coordinated to address these challenges through cultural adaptation, target text rewriting, revision, and conditional visual adjustment. 
The framework strengthens target text adaptation with coordinated feedback to handle difficult cases that require deeper cultural or visual intervention. 
We evaluate the framework on bidirectional Chinese-English meme transcreation using both human evaluation and LLM-as-a-Judge. 
Our method consistently outperforms all baselines across both human evaluation and LLM-as-a-Judge settings. In human evaluation, it achieves the best performance on all four dimensions and delivers a 33.1\% average improvement over the strongest baseline, while in LLM-as-a-Judge, it attains the highest Top-1 ranking rate (60
\% vs. 26\% for the second-best baseline). Further analysis indicates that each component contributes to the performance, highlighting the effectiveness of the proposed architecture, and our error analysis suggests that the remaining bottlenecks lie in humor reconstruction and image-text alignment rather than simple cultural knowledge gaps, pointing to future work on humor transfer. \footnote{The code is available at \url{https://github.com/Jingyi62/TransMeme}.}
\end{abstract}






\maketitle
% \vspace{-4pt}
\section{Introduction}

Internet memes have become a pervasive form of online communication, serving as compact carriers of humor, stance, emotion, and social commentary \cite{shifman2013memes}. 
\begin{figure*}[t]
  \centering
  \includegraphics[width=\linewidth]{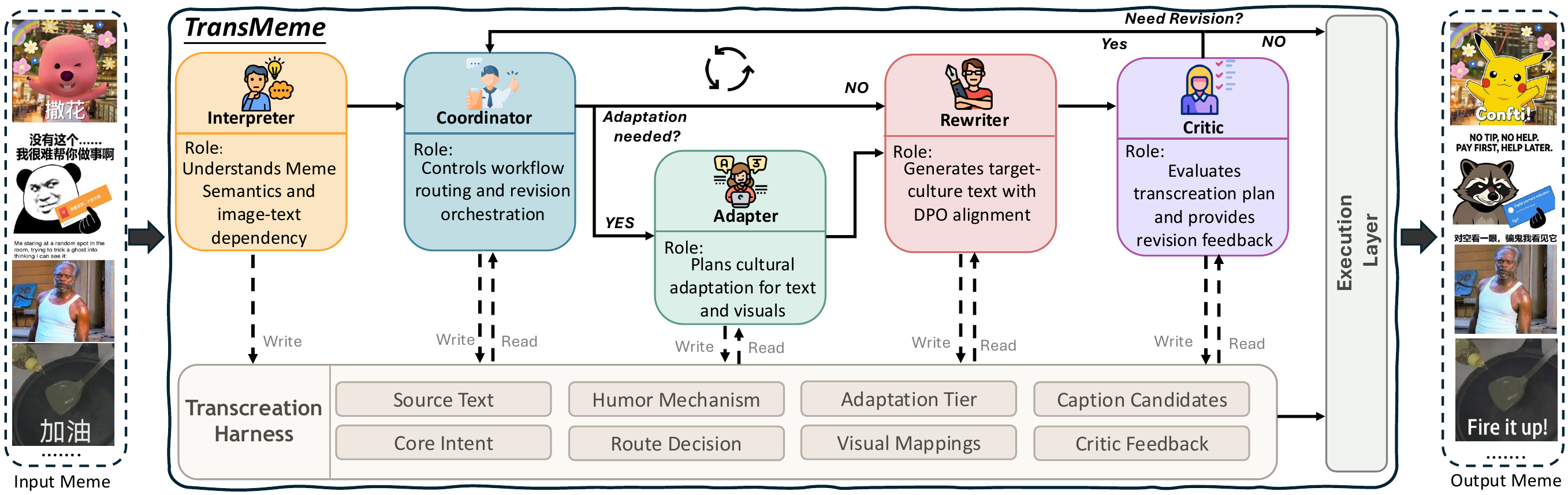}
  \caption{
  Overview of our multi-agent framework for cross-cultural meme transcreation.}
  \label{fig:method_overview}
\end{figure*}
As a widely shared form of digital expression, they also shape how people participate in online culture \cite{shifman2014memes}, express political and social positions \cite{ahmed2024breaking}, and negotiate identity across communities. 
This makes cross-cultural meme transfer important for both multilingual digital communication and understanding how online information travels across linguistic and cultural boundaries. 
Yet such a transfer is far from straightforward. 
Unlike ordinary text, memes derive their effect from the interaction between short written text, visual templates, and shared cultural knowledge \cite{shifman2014memes,hakokongas2020persuasion}, and their meaning is often implicit, relying on background assumptions, discourse conventions, symbolic imagery, or locally recognizable references rather than literal wording alone \cite{hakokongas2020persuasion}. 
A successful target meme needs to do more than restate the source content in another language: it needs to reconstruct a communicative effect that works for a different audience.

This task is related but not reducible to several similar problems. 
Specifically, it shares cross-lingual transfer with machine translation, visual grounding with multimodal translation, rhetorical reformulation with text rewriting, and visual modification with image transcreation \cite{bahdanau2015neural,specia-etal-2016-shared,krishna-etal-2020-reformulating,khanuja-etal-2024-image}. Recent work has introduced cross-cultural meme transcreation as an end-to-end benchmarked task \cite{zhao2026beyond}. 
However, it does not provide sufficiently strong empirical performance or the necessary analysis to address the problem. 
So we begin with an analysis, which suggesting that meme transcreation is shaped by three challenges: \textit{Culture-specific knowledge understanding}, because meme meaning often relies on local knowledge, conventions, and socially shared references; \textit{Intent and Tone Preservation}, because humor, sarcasm, ridicule, and stance often need to be re-expressed rather than translated literally; and \textit{Multimodal Consistency}, because text and image jointly construct the meme effect, so adapting one side may weaken or break its coherence with the other. 
This analysis shows that cross-cultural meme transcreation is not simply a translation, but a challenging adaptation problem governed by constraints on cultural understanding, intent preservation, and multimodal consistency.

To address these challenges, we propose \textit{TransMeme}, a multi-agent framework for cross-cultural meme transcreation that decomposes the task into stages following the structure revealed by our analysis.
The \emph{Adapter} addresses culture-specific knowledge understanding through explicit planning, deciding whether source references should be preserved, reformulated, or remapped for the target audience. 
The \emph{Rewriter}, further strengthened with \emph{DPO}, handles intent and tone preservation by generating target text that better preserves humor, stance, and communicative force under explicit adaptation constraints. 
The \emph{Coordinator} and \emph{Critic} handle multimodal consistency by routing difficult cases to deeper adaptation, checking whether the adapted text remains compatible with the visual template, and triggering targeted revision when cultural, semantic, or visual inconsistencies arise. 
Together, these components enable \textit{TransMeme} to preserve transferable elements, rewrite what needs adaptation, and revise difficult cases when necessary.

\begin{table}[t]
\centering
\small
\setlength{\tabcolsep}{4.5pt}
 \caption{Comparison between meme transcreation and related tasks. \textbf{XLang}: cross-lingual transfer; \textbf{Text}: textual adaptation; \textbf{Visual}: visual understanding or editing; \textbf{Culture}: cultural adaptation; \textbf{I-T}: image-text coherence.}
\vspace{-1em}
\begin{tabular}{lccccc}
\toprule
\textbf{Task} & \textbf{XLang} & \textbf{Text} & \textbf{Visual} & \textbf{Culture} & \textbf{I-T} \\
\midrule
Machine Translation    & \cmark & \cmark & \xmark & \xmark & \xmark \\
Multimodal Translation & \cmark & \cmark & \cmark & \xmark & \xmark \\
Text Rewriting         & \xmark & \cmark & \xmark & \xmark & \xmark \\
Image Editing          & \xmark & \xmark & \cmark & \xmark & \xmark \\
Meme Transcreation     & \cmark & \cmark & \cmark & \cmark & \cmark \\
\bottomrule
\end{tabular}
\label{tab:task_comparison}
\vspace{-1em}
\end{table}
We evaluate \textit{TransMeme} on bidirectional Chinese-English meme transcreation using both human and LLM evaluation. 
The results show that \textit{TransMeme} consistently outperforms all baselines. 
On human evaluation, it achieves the best scores on all four dimensions, with an overall average of 4.122, surpassing the strongest baseline at 3.097. 
The same advantage remains clear on the full benchmark, where our method is selected as the top-ranked system most frequently, reaching a top-1 rate of 0.600.
We further conduct component analysis to examine where these gains come from. 
The DPO-enhanced \emph{Rewriter} improves target text quality, raising the average text-level score from 2.839 without DPO to 4.180.
The \emph{Critic} based revision loop is effective: among the 125 cases that enter revision, the revised result is preferred in 105 cases. 
Explicit cultural planning proves similarly crucial. 
On the 181 samples that require this module, the full system is preferred over the version without cultural planning in 180 cases. 
Together, these results show that the gains of \textit{TransMeme} come not merely from the prompt but from challenge-aligned adaptation, revision, and control.
Our contributions are threefold:
\begin{itemize}[leftmargin=*]
    \item We provide an explicit task analysis of cross-cultural meme transcreation, clarifying its core requirements and identifying three central challenges: Culture-specific knowledge understanding, Intent and Tone Preservation, and Multimodal Consistency.
    \item We propose \textit{TransMeme}, a challenge-driven multi-agent framework that integrates structured understanding, cultural planning, text rewriting with a \emph{DPO}-enhanced rewriter, critic-guided revision, and conditional visual execution.
    
    \item We evaluate \textit{TransMeme} through human and LLM-based evaluation, and component analyses, showing that adaptation planning and targeted revision yield consistent gains over strong one-shot and structured baselines.
\end{itemize}

\section{Task Analysis}
\label{sec:task}

\subsection{Task Requirements}

Given a source meme, the task is to produce a target-culture meme for different audiences. 
The output may preserve the original image, or edit it, while the accompanying text may need to be rewritten rather than directly translated.
This task is related to several established tasks (see \Cref{tab:task_comparison}). 
It involves cross-lingual transfer, as in machine translation \citep{bahdanau2015neural}, and may also use visual information, as in multimodal translation \citep{specia-etal-2016-shared}. 
It requires textual reformulation, which connects it to text rewriting \citep{krishna-etal-2020-reformulating}. 
It can further involve visual adaptation, which is also studied in image transcreation \citep{khanuja-etal-2024-image}. 
At the same time, cross-cultural meme transcreation is not reducible to any of these tasks in isolation.
We define three requirements specific to meme transcreation.
First, \textbf{intent preservation} requires the target meme to retain the source meme's core communicative effect. 
This includes humor, sarcasm, ridicule, irony, emotional stance, and other pragmatic functions.
Second, \textbf{cultural adaptation} requires the result to be understandable, natural, and appropriate for the target audience. 
A transcreated meme should work within the target culture rather than merely restating the source content in another language.
Third, \textbf{image-text coherence} requires the adapted text and the final image to support the same meme effect. 
The meme should function as a unified multimodal artifact rather than as a plausible sentence placed on an unrelated picture. 
This requirement is especially important for memes, since prior work has shown that meme meaning and rhetoric emerge from the interaction of verbal and visual elements rather than from either modality alone \citep{hakokongas2020persuasion,zhao2026beyond}.

\subsection{Core Challenges}

These requirements lead to three challenge categories. 
Each challenge corresponds to one major requirement of the task, although they often appear together in practice.

\mypara{Culture-specific knowledge understanding}
The requirement of cultural adaptation creates this
challenge. 
Many memes rely on source culture entities, stereotypes, public events, discourse conventions, or shared background knowledge. 
In such cases, the text alone conveys only part of the meme's meaning.
A literal transfer may preserve wording but fail for the target audience. 
The system must decide whether to keep the original reference, replace it with a target-side analogue, or reframe it. 
This makes cultural adaptation a decision problem rather than a simple translation.

\mypara{Intent and Tone Preservation}
The requirement of intent preservation creates the second challenge. 
Memes often depend on sarcasm, irony, exaggeration, ridicule, or stance. 
These effects are rarely preserved by semantic transfer alone. 
The text can remain close in meaning and still lose its punch, tone, or humor. 
The system needs to reconstruct communicative effect, not just propositional content. 
This is why text adaptation in meme transcreation is closer to pragmatic rewriting than to ordinary translation.

\mypara{Multimodal Consistency}
The requirement of image-text coherence creates the third challenge. 
In memes, text and image usually work together to produce the final effect. 
The same text may succeed with one template and fail with another. 
The same image may support one rhetorical framing but conflict with a different one. 
Thus, local adaptation at the text level is not enough. 
The system must also judge whether the original image remains suitable, whether it should be edited, or whether a new realization is needed.

These challenges are tightly connected. 
A meme may contain a culture-specific reference, express that reference through sarcasm, and rely on a particular expression at the same time. 
For this reason, cross-cultural meme transcreation should not be treated as a simple generation task. 
It requires explicit understanding of meme meaning, deliberate planning of cultural adaptation, careful rewriting of the text, and a final check on multimodal coherence.

\section{Method}

\subsection{System Overview}
We propose a multi-agent framework for cross-cultural meme transcreation that takes a source meme image as input and generates a target meme.
As shown in Figure~\ref{fig:method_overview}, the framework consists of five agents-the \emph{Interpreter}, \emph{Coordinator}, \emph{Adapter}, \emph{Rewriter}, and \emph{Critic}-followed by the \emph{Execution Layer} for final visual realization.
The design directly follows the three challenge categories in Section~\ref{sec:task}. 
The \emph{Adapter} handles the understanding of culture-specific knowledge through explicit adaptation planning. 
The \emph{Rewriter}, which is further optimized with DPO for meme text adaptation, handles intent and tone preservation through target text rewriting.
The \emph{Coordinator} and \emph{Critic} address Multimodal Consistency by checking whether the adapted text fits the visual template and by triggering revision when needed.
The \emph{Interpreter} provides the source meme analysis for subsequent decisions, while the \emph{Transcreation Harness} maintains the intermediate pipeline state and supports execution.

\subsection{Interpreter and Coordinator: Understanding and Control}
The framework begins with source meme understanding and workflow control. 
The \emph{Interpreter} recovers the source meme's communicative effect beyond surface text and writes the result into the \emph{Transcreation Harness}. 
Its analysis has two parts. First, it infers the meme's semantic and pragmatic content, including core intent, humor mechanism, emotional stance, relevant background knowledge, and source side elements that should ideally be preserved. 
Second, it assesses image and text dependency, including whether the intended effect relies on visual entities, facial expression, symbolic cues, or scene logic, and whether visual elements belong to a fixed template structure or editable content.

Based on this analysis, the \emph{Coordinator} controls how the pipeline proceeds. 
For simple cases, it may choose a lightweight path with limited rewriting. 
For difficult cases, those involving stronger cultural dependency, rhetorical reframing, or tight image and text coupling, it routes the instance to the full transcreation path. 
The routing decision is guided by cues such as literal traps, text length, reaction meme patterns, and coupling strength. 
After later stages produce candidate outputs, the \emph{Coordinator} also manages revision based on critical feedback, deciding whether to accept the current result, trigger another round of rewriting, return the case for cultural remapping, or adjust the visual execution plan.

\subsection{Adapter: Cultural Adaptation Planning}
The \emph{Adapter} constructs an explicit transcreation plan before text generation. 
Its role is to convert the source-side understanding into a target-side adaptation strategy that specifies how the meme should be localized for the target culture while preserving the source intent.
This stage is designed to address the challenge of \textbf{Culture-specific knowledge understanding}.
It explicitly identifies and adapts culturally specific elements that may not transfer effectively, including named entities, social references, discourse conventions, and stereotypes.
It then estimates the likely comprehension gap for the target audience and determines whether these elements can be preserved, lightly reformulated, or replaced with target-culture counterparts.
A decision of this stage is the adaptation tier, defined as one of three levels: literal, minimal rephrase, and cultural map. 
These tiers indicate whether the meme is largely portable, requires constrained rewriting, or needs deeper cultural remapping.
In this way, the \emph{Adapter} operationalizes cultural adaptation as an explicit decision layer before generation.
The \emph{Adapter} also produces a visual adaptation plan.
It specifies whether visual cultural adaptation is needed, which visual elements should remain fixed, and which parts may require replacement or reinterpretation.
Although its primary role is to solve Culture-specific knowledge understanding, this visual planning step also supports \textbf{Multimodal Consistency}, since later text adaptation must remain compatible with the final visual realization.
The resulting plan is stored in the \emph{Transcreation Harness} and guides both later rewriting and final execution.

\begin{table}[t]
\centering
\small
\setlength{\tabcolsep}{3.2pt}
\caption{Alignment between human and LLM evaluation.}
\vspace{-1em}
\begin{tabular}{lcccccc}
\toprule
\multirow{2}{*}{\textbf{Metric}} &
\multicolumn{2}{c}{\textbf{All}} &
\multicolumn{2}{c}{\textbf{ZH$\rightarrow$EN}} &
\multicolumn{2}{c}{\textbf{EN$\rightarrow$ZH}} \\
\cmidrule(lr){2-3} \cmidrule(lr){4-5} \cmidrule(lr){6-7}
& \textbf{Stat.} & \textbf{MAE} & \textbf{Stat.} & \textbf{MAE} & \textbf{Stat.} & \textbf{MAE} \\
\midrule
Intent ($\rho$) & 0.802 & 0.617 & 0.811 & 0.592 & 0.787 & 0.642 \\
Adaptation ($\rho$) & 0.693 & 0.791 & 0.710 & 0.791 & 0.686 & 0.792 \\
Coherence ($\rho$) & 0.783 & 0.680 & 0.778 & 0.685 & 0.791 & 0.676 \\
Quality ($\rho$) & 0.789 & 0.748 & 0.807 & 0.669 & 0.765 & 0.827 \\
\midrule
Ranking ($\tau$) & 0.422 & - & 0.419 & - & 0.425 & - \\
Ranking (Spearman) & 0.507 & - & 0.507 & - & 0.508 & - \\
Top-1 agreement & 0.890 & - & 0.870 & - & 0.910 & - \\
\bottomrule
\end{tabular}
\vspace{-1em}
\label{tab:llm_alignment_single}
\end{table}

\subsection{Rewriter: Text Realization under Adaptation Constraints}
The \emph{Rewriter} generates target-side meme text candidates conditioned on the current \emph{Transcreation Harness}. This stage addresses the challenge of \textbf{Intent and Tone Preservation} by reconstructing humor, sarcasm, irony, ridicule, and stance in a form that remains effective for the target audience. Rather than performing unrestricted generation, the \emph{Rewriter} is guided by the explicit adaptation constraints produced by earlier stages, including the adaptation tier, cultural mappings, rhetorical goals, and visual constraints, so that the output remains aligned with the overall transcreation plan.
We further strengthen this stage through \textbf{Direct Preference Optimization (DPO)}. 
Because meme transcreation requires specialized rewriting preferences, we build a transcreation-oriented preference dataset from colloquial expressions and slang collected from online sources, and obtain human rankings over multiple rewritten candidates. 
Given a structured transcreation context \(x\), we denote the preferred rewrite set as \(Y^{+}\) and the rejected rewrite set as \(Y^{-}\). 
Pairwise preferences are then constructed by sampling \((y^{+}, y^{-})\) with \(y^{+} \in Y^{+}\) and \(y^{-} \in Y^{-}\). 
The \emph{Rewriter} is optimized with the standard DPO objective:
\[
\mathcal{L}_{\mathrm{DPO}} = - \log \sigma \Bigg( \beta \Bigg[
\log \frac{\pi_{\theta}(y^{+}\mid x)}{\pi_{\mathrm{ref}}(y^{+}\mid x)}
-
\log \frac{\pi_{\theta}(y^{-}\mid x)}{\pi_{\mathrm{ref}}(y^{-}\mid x)}
\Bigg] \Bigg).
\]

Here, \(x\) includes the source meme text together with the adaptation constraints.
This alignment encourages the model to prefer rewrites that better preserve source intent, better fit the target cultural context, and better match the planned visual setting.
The generated text candidates are written back into the \emph{Transcreation Harness} for subsequent evaluation and possible revision.

\subsection{Critic: Evaluation and Revision Feedback}
The \emph{Critic} evaluates intermediate outputs during transcreation and provides revision feedback when needed.
It assesses whether the current result satisfies the three task requirements in Section~\ref{sec:task}, including intent preservation, cultural adaptation, and image text coherence.
This stage is most directly related to the challenge of \textbf{Multimodal Consistency}, since it checks whether the adapted text, cultural mapping, and visual plan still work together as a coherent meme.
Concretely, the \emph{Critic} examines generated text, proposed cultural mappings, and visual adaptation plans.
It checks whether the current output preserves the source meme's communicative effect, whether the cultural adaptation is appropriate for the target audience, and whether the adapted text remains consistent with the visual realization.
When problems are identified, they produce targeted feedback that specifies which part requires revision, such as text rewriting, cultural remapping, or visual adjustment.
The critique results are written into the \emph{Transcreation Harness} and used in the subsequent revision process before final execution.

\begin{figure*}[t]
    \centering
    \vspace{-1em}
    \includegraphics[width=\textwidth]{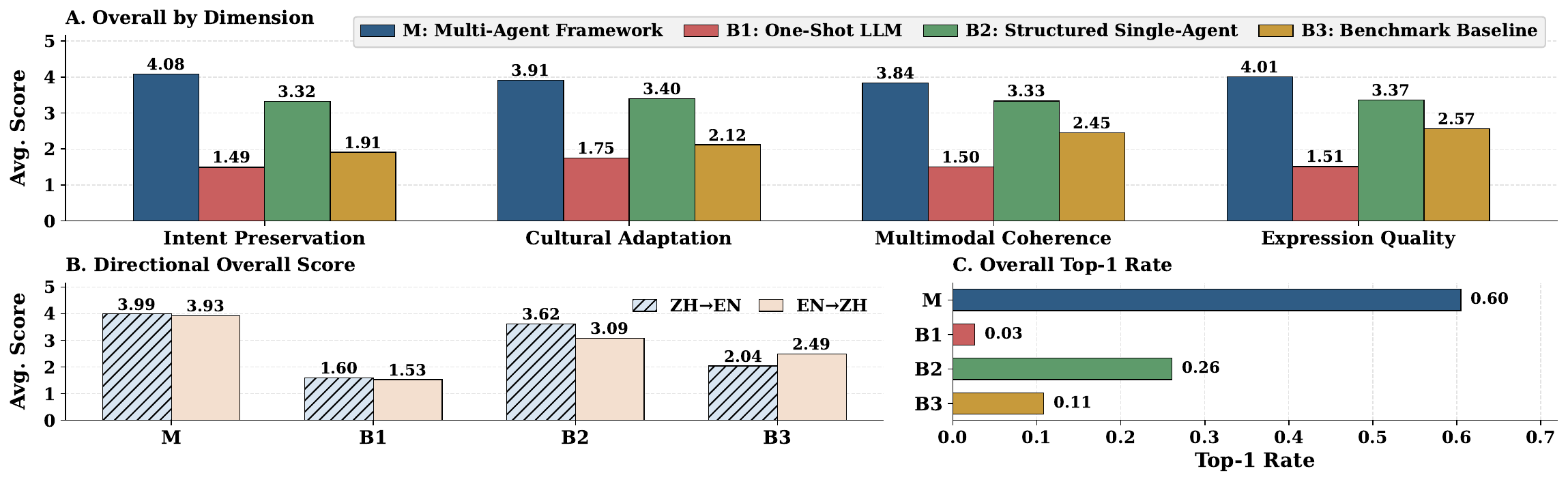}
    \vspace{-2.5em}
    \caption{Main LLM-based results: (A) Dimension-wise comparison. (B) Average-score by transcreation direction. (C) Top-1 selection rate. Our method performs best overall, remains robust in both directions, and is most frequently ranked first.}
    \vspace{-1.5em}
    \label{fig:main_results_combined}
\end{figure*}

\subsection{Transcreation Harness and Execution Layer}
The \emph{Transcreation Harness} maintains the intermediate state of the pipeline.
It stores intermediate outputs and signals from different stages, including but not limited to source meme understanding, routing decisions, cultural mappings, generated text, and critic feedback.
In this way, it preserves the accumulated decisions made during transcreation and provides the runtime structure for the full framework.
After revision is complete, the finalized state is passed to the \emph{Execution Layer} for final visual realization.
The \emph{Execution Layer} applies the approved target side text together with the selected visual strategy, which may preserve the original image, replace the text, or perform stronger visual editing when deeper adaptation is required.
The final target meme is therefore rendered from the structured decisions produced by the full pipeline.

\section{Experiments}
\label{sec:exp}

\subsection{Experimental Settings}

\subsubsection{Dataset, Baselines, and Implementation Details}
Following \cite{zhao2026beyond}, we evaluate bidirectional Chinese--English meme transcreation on 1,000 randomly sampled memes, including 500 ZH$\rightarrow$EN and 500 EN$\rightarrow$ZH instances.
Under identical input and output settings, we compare \textbf{M}, our multi-agent framework \textit{TransMeme}, with three baselines. 
\textbf{B1} (\textit{Direct One-Shot Transcreation}) uses \texttt{Qwen-image-2.0} to generate the target meme in a single pass. 
\textbf{B2} (\textit{Structured Single-Agent Transcreation}) uses \texttt{Qwen3-VL-Plus} with a prompt covering meme understanding, cultural adaptation, text rewriting, and image--text coherence. The resulting structured transcreation plan is then executed by \texttt{Qwen-image-2.0}.
\textbf{B3} (\textit{Reference Benchmark Baseline}) is adopted from \cite{zhao2026beyond}. 
For \textit{TransMeme}, the Interpreter uses \texttt{Qwen3-VL-Plus}; the Coordinator, Adapter, and Critic use \texttt{Qwen3-Max}; the Rewriter is initialized from \texttt{Qwen3-8B}; and the Execution Layer uses \texttt{Qwen-Image-2.0}. 
The Rewriter is trained with DPO using TRL and LoRA. To construct the training corpus, we collect web-sourced colloquial expressions and slang in Chinese and English. 
For each source instance, \texttt{Qwen3-8B} generates multiple translation candidates, which are ranked by human annotators, yielding a bilingual ranked corpus containing 500 Chinese and 500 English instances. 
A single bidirectional Rewriter is trained on the merged data. 
The DPO corpus is constructed independently of the meme benchmark and contains no source text appearing in the evaluation meme images. 
At inference time, the Rewriter generates three candidate texts, after which the Critic scores them, selects the best candidate, and provides revision feedback when necessary.

\subsubsection{Human Evaluation Protocol}

We conduct human evaluation because cross-cultural meme transcreation requires jointly assessing semantic fidelity, cultural appropriateness, multimodal compatibility, and the overall quality of the final presentation. 
We randomly sample 200 instances (100 ZH$\rightarrow$EN and 100 EN$\rightarrow$ZH). 
Our evaluation dimensions follow the Section~\ref{sec:task}: \textit{Intent Preservation}, \textit{Cultural Adaptation}, and \textit{Multimodal Coherence} correspond to the three task requirements, while \textit{Expression Quality} measures fluency, naturalness, and the overall presentation quality of the final meme in the target culture. 
Each sample is rated on a 5-point Likert scale on all dimensions, and annotators also provide a full ranking over the compared systems for each source meme.
We employ five bilingual annotators familiar with both meme cultures. 
Before annotation, all annotators receive training based on a guideline covering the evaluation dimensions and common edge cases. The full guideline is included in the supplementary material.
All annotators independently evaluate every sample using the original meme and anonymized system outputs presented in randomized order.
For scalar judgments, each system output for a given source meme is treated as one evaluation item. For each dimension, we measure agreement among the five annotators using ICC(2,1), which evaluates absolute agreement on single ratings under a two-way random-effects model \cite{shrout1979intraclass,koo2016guideline}.
For ranking-based judgments, we measure agreement with Kendall's $W$, which captures concordance over the full rankings \cite{kendall1939problem}. 
To obtain a reference human ranking for later comparison, we average per-system ranks across annotators within each sample and derive the corresponding human top-1 label from the aggregated ranking. 
Winner agreement among human annotators is computed as the mean pairwise Cohen's $\kappa$ over their top-1 selections, where $\kappa$ measures agreement beyond chance for categorical decisions \cite{cohen1960coefficient}. 
Overall, the human judgments exhibit reliable agreement. 
For scalar ratings, ICC(2,1) is 0.842 for intent, 0.772 for adaptation, 0.821 for coherence, and 0.809 for quality. 
Agreement is slightly higher for ZH$\rightarrow$EN than for EN$\rightarrow$ZH across all four dimensions: 0.858 versus 0.824 for intent, 0.797 versus 0.746 for adaptation, 0.832 versus 0.809 for coherence, and 0.832 versus 0.784 for quality. 
For comparative judgments, full-ranking consistency reaches a Kendall's $W$ of 0.604 overall, with similar values for ZH$\rightarrow$EN ($W=0.606$) and EN$\rightarrow$ZH ($W=0.601$). Winner agreement is moderate, with an overall Cohen's $\kappa$ of 0.554, and is slightly higher for ZH$\rightarrow$EN ($\kappa=0.572$) than for EN$\rightarrow$ZH ($\kappa=0.538$).

\begin{figure*}[t]
  \centering
  \vspace{-1em}
  \includegraphics[width=\linewidth]{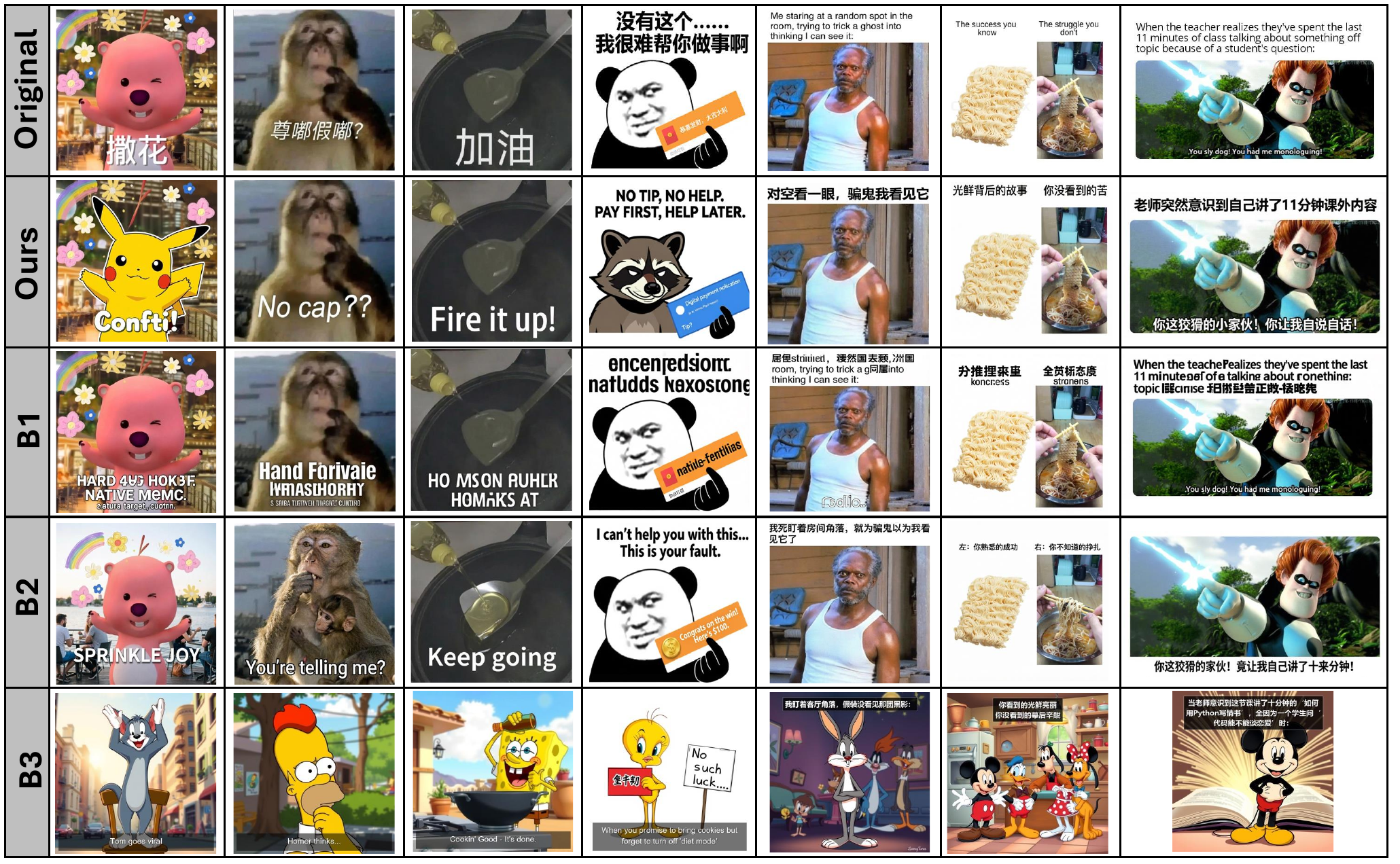}
  \vspace{-2em}
  \caption{
  Qualitative comparison of meme transcreation results. Our method produces target memes that better preserve humor, improve cultural appropriateness, and maintain image-text coherence.}
  \vspace{-1.5em}
  \label{fig:use_case}
\end{figure*}
\subsubsection{LLM Evaluator Validation}

To extend evaluation beyond the human-annotated subset, we use an LLM-based evaluator (\texttt{Qwen3.5-plus}) for large-scale comparison on the full evaluation set. 
Following the same guidelines as human judges, it takes the original meme, the final rendered outputs of all systems as input, and outputs the same four scalar scores together with an overall ranking. 
We validate the evaluator on the 200-sample human-annotated subset. 
For scalar scores, we compare each LLM score with the mean human rating for the corresponding source meme-system pair using Spearman correlation and MAE. 
For ranking, we compare the LLM ranking with the aggregated human ranking for each sample using Kendall's $\tau$ and Spearman correlation, averaged over all samples, and also report LLM-human Top-1 agreement against the aggregated human top-1 label. 
As shown in Table~\ref{tab:llm_alignment_single}, the evaluator shows good agreement with aggregated human judgments on scalar scores, with overall Spearman correlations ranging from 0.693 to 0.802 across the four dimensions. 
Alignment is strongest on \textit{Intent Preservation} and weakest on \textit{Cultural Adaptation}, the most culturally sensitive dimension. 
For system-level comparison, the evaluator achieves 89.0\% Top-1 agreement. 
Full-ranking alignment is more moderate, with mean Kendall's $\tau = 0.422$ and mean Spearman correlation $= 0.507$, but remains stable across both directions. 
These results support using the LLM evaluator as a scalable proxy for comparison, while human evaluation on the subset remains the primary validity reference.

\subsection{Main Results}

We report the main results from two perspectives, with Table~\ref{tab:main_results_human} showing aggregated human evaluation on the subset and Figure~\ref{fig:main_results_combined} summarizing LLM-based evaluation on the full set. 
Across both views, \textit{TransMeme} consistently outperforms all baselines.
As shown in Table~\ref{tab:main_results_human}, \textbf{M} achieves the best human evaluation scores on all dimensions, reaching 3.950 on Intent Preservation, 4.097 on Cultural Adaptation, 4.111 on Multimodal Coherence, and 4.328 on Expression Quality, with an overall average of 4.122. 
\textbf{B2} is the strongest baseline with an average of 3.097, while \textbf{B3} and \textbf{B1} trail substantially at 1.811 and 1.217, respectively. 
Moreover, the gains of \textbf{M} over all baselines are statistically significant on every dimension and on the overall average score under two-sided Wilcoxon signed-rank tests (\textit{all} $p<0.001$), and remain significant after Bonferroni correction.
Figure~\ref{fig:main_results_combined} shows that the same pattern holds at scale on the full evaluation set. 
In the overall dimension-wise comparison, \textbf{M} consistently outperforms all baselines on Intent Preservation, Cultural Adaptation, Multimodal Coherence, and Expression Quality. 
The gap over \textbf{B2}, the strongest baseline, remains clear across all dimensions, indicating that the gains do not come merely from adding partial structure to prompting, but from explicitly decomposing transcreation into understanding, cultural adaptation planning, rewriting, and revision. 
The directional analysis in Figure~\ref{fig:main_results_combined}(B) further shows that \textbf{M} remains the best-performing method in both ZH$\rightarrow$EN and EN$\rightarrow$ZH settings. 
The relative ordering of methods is stable across directions, suggesting that the framework is robust rather than overfitted to a single transfer direction. 
M, B1, and B2 perform better in the ZH$\rightarrow$EN setting, whereas B3 shows the opposite pattern. 
Finally, Figure~\ref{fig:main_results_combined}(C) highlights the ranking advantage of \textbf{M}. 
Our method is selected as the top-ranked system more often than others, indicating that it is not only better on average, but also more frequently preferred as the best output on individual examples. 
Taken together, the human and LLM results consistently confirm the effectiveness of our framework.
As shown in Figure~\ref{fig:use_case}, we further present a randomly selected set of examples for qualitative analysis. These cases are consistent with the quantitative results above and help illustrate the behaviors underlying the performance differences. 
Specifically, they show that the advantage of \textbf{M} lies in more natural humor reconstruction, more effective cultural substitution when source-specific references cannot be transferred directly, and stronger preservation of image-text coherence. 
By contrast, the baselines often produce literal or awkward target text, insufficient cultural adaptation, or mismatches between the image and the text.

\begin{table}[t]
\centering
\small
\caption{Aggregated human evaluation results, where our method performs best across all dimensions. Avg. denotes the mean of the four dimension scores.}
\vspace{-1em}
\begin{tabular}{lccccc}
\toprule
Method & Intent & Adaptation & Coherence & Quality & Avg. \\
\midrule
M  & \textbf{3.950} & \textbf{4.097} & \textbf{4.111} & \textbf{4.328} & \textbf{4.122} \\
B1 & 1.187 & 1.328 & 1.180 & 1.172 & 1.217 \\
B2 & 3.385 & 3.138 & 3.380 & 2.484 & 3.097 \\
B3 & 1.716 & 1.818 & 1.910 & 1.800 & 1.811 \\
\bottomrule
\end{tabular}
\vspace{-1em}
\label{tab:main_results_human}
\end{table}
\subsection{Component Analysis}

To better understand where the gains of \textit{TransMeme} come from, we analyze three key components separately: the rewriter design, the critic-revision loop, and the cultural planning module. 

\subsubsection{Rewriter Quality Analysis}

\begin{figure}[t]
    \centering
    \vspace{-1em}
    \includegraphics[width=\columnwidth]{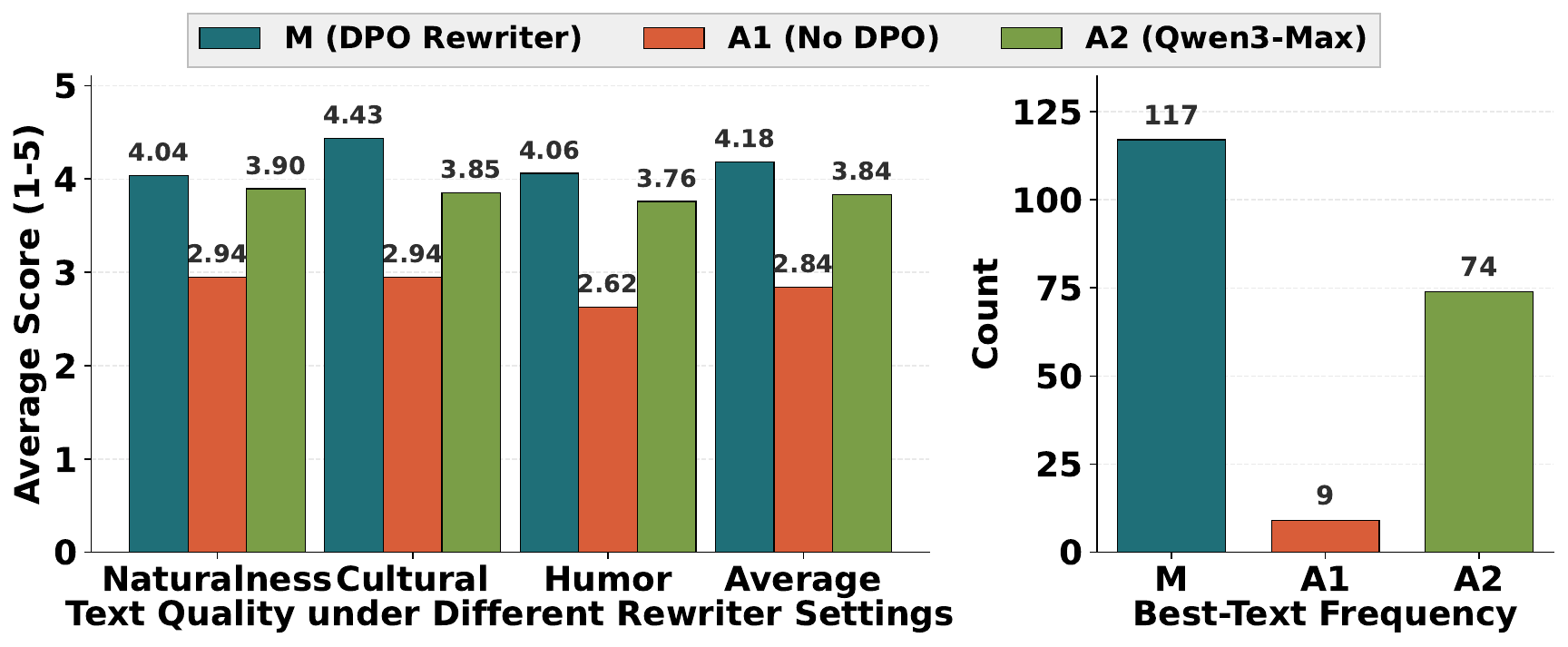}
    \vspace{-2em}
    \caption{Rewriter quality analysis: M consistently outperforms A1 and A2, confirming the benefit of DPO. }
    \vspace{-1em}
    \label{fig:rewriter_quality}
\end{figure}

We compare the full model (\textbf{M}) against two variants: \textbf{A1}, which removes DPO while keeping the rest of the pipeline unchanged, and \textbf{A2}, which replaces the final rewriter with \texttt{Qwen3-Max}. 
Since these variants only affect text generation, we evaluate them using text-level judgments rather than scores. 
We randomly sample 200 instances, including 100 Chinese-to-English and 100 English-to-Chinese cases, and use \texttt{Qwen3-Max} as the evaluator to score generated texts along three dimensions: Naturalness, Cultural quality, and Humor. 
We include \textbf{A2} as a strong off-the-shelf baseline: compared with our DPO-trained \texttt{Qwen3-8B} rewriter, \texttt{Qwen3-Max} is a substantially larger general-purpose model.
As shown in Figure~\ref{fig:rewriter_quality}, \textbf{M} consistently achieves the best performance across all dimensions, obtaining 4.035 on Naturalness, 4.435 on Cultural quality, and 4.060 on Humor, with an overall average of 4.177. 
Removing DPO leads to a substantial drop: \textbf{A1} decreases to 2.945 on Naturalness, 2.945 on Cultural quality, and 2.625 on Humor, with an average of 2.838. 
\textbf{A2}, despite serving as a strong replacement baseline, remains below the full model, reaching 3.895 on Naturalness, 3.850 on Cultural quality, and 3.760 on Humor, with an average of 3.835. 
The same pattern is reflected in the best-method comparison in Figure~\ref{fig:rewriter_quality} (right), where \textbf{M} is selected as the best text 117 times, compared with 74 for \textbf{A2} and 9 for \textbf{A1}. 
These results show that DPO contributes substantially to text quality.

\subsubsection{Critic Loop Analysis}

We analyze the critic loop in terms of trigger frequency, error types, and revision effectiveness. On the full evaluation set, 125 of 1,000 samples enter the loop, corresponding to a trigger rate of 12.5\%. Among these cases, the average loop count is 1.72 and the maximum is 2, with 35 samples revised once and 90 revised twice. The trigger rate differs substantially by direction: 106 of 500 ZH$\rightarrow$EN cases (21.2\%) enter the loop, compared with only 19 of 500 EN$\rightarrow$ZH cases (3.8\%), suggesting that ZH$\rightarrow$EN requires more iterative cultural and pragmatic adaptation. Of the 125 triggered cases, 75 (60.0\%) involve \textit{cultural mismatch}, 47 (37.6\%) involve \textit{intent drift}, 2 (1.6\%) involve image--text inconsistency, and 1 (0.8\%) involves weak humor or awkward wording. For ZH$\rightarrow$EN, cultural mismatch accounts for 62.3\% of cases and intent drift for 35.8\%; for EN$\rightarrow$ZH, the two categories each account for 47.4\%, indicating that the critic mainly repairs cultural and semantic failures rather than local stylistic issues. To evaluate revision effectiveness, we compare the pre- and post-revision outputs of the 125 triggered samples using \texttt{Qwen3.5-397b} as the evaluator. The revised outputs are preferred in 105 cases (84.0\%), while the pre-revision outputs are preferred in 20 cases (16.0\%). Revision improves Intent Preservation from 2.352 to 3.632, Cultural Adaptation from 2.272 to 3.704, and Expression Quality from 2.736 to 3.792, raising the overall average from 2.453 to 3.709. The average score also increases from 2.396 to 3.629 for ZH$\rightarrow$EN and from 2.772 to 4.158 for EN$\rightarrow$ZH, demonstrating that the critic loop effectively corrects culturally and semantically problematic generations.

\subsubsection{Cultural Planning Analysis}

We examine the role of the cultural planning module, which is triggered for 181 of 1,000 samples, including 155 ZH$\rightarrow$EN cases and 26 EN$\rightarrow$ZH cases.
This clear asymmetry suggests that explicit cultural planning is more frequently required when adapting Chinese memes for English audiences. 
Representative cases illustrate the types of textual and visual adaptation handled by the module. 
In an EN$\rightarrow$ZH meme built around ``\textit{celebrating my 18th birthday}'' at a bar, the humor depends on the legal drinking age and the irony of being an underage regular, requiring cultural reinterpretation rather than literal translation. 
In a ZH$\rightarrow$EN meme containing the Chinese phrase meaning ``\textit{My period's here}'', the euphemism is correctly recognized as referring to menstruation rather than a literal aunt, preserving its intended tone and pragmatic force. The module also supports culturally appropriate visual substitution.
As shown in Figure~\ref{fig:use_case}, the first-column example replaces Loopy with Pikachu as a more recognizable and accessible expressive figure for the target audience, while the fourth-column example replaces a red envelope with a transfer record, substituting a culture-specific social symbol with a more immediately interpretable equivalent. 
To quantify the module's contribution, we compare the full system with a variant that skips cultural planning while keeping all other components unchanged. 
The comparison is conducted on the 181 triggered samples using \texttt{Qwen3.5-397b} as the evaluator across Intent Preservation, Cultural Adaptation, and Expression Quality.
The full system is preferred in 180 cases, whereas the variant without cultural planning is preferred in only 1 case. 
It also achieves substantially higher average scores of 3.972, 3.972, and 3.983 on the three dimensions, respectively, compared with 1.600, 1.511, and 1.606 for the variant. 
These results show that cultural planning is not only frequently activated but also essential in cases requiring non-literal cultural reinterpretation or visual substitution.

\subsection{Error Analysis}

Despite its overall advantage, our framework does not rank first on 400 of the 1,000 evaluation samples. 
A manual analysis of these non-top-ranked cases identifies three main error types: weak humor reconstruction (271/400, 67.8\%), image--text mismatch (108/400, 27.0\%), and cultural knowledge gaps (21/400, 5.3\%). 
Weak humor reconstruction is therefore the dominant failure mode, accounting for more than two-thirds of all non-top-ranked cases.
The most common pattern is that the system preserves the source intent but fails to recreate an equally effective humorous effect in the target culture. 
These outputs are usually understandable and semantically acceptable, yet they are often less witty, less natural, or less punchy than the preferred alternative. 
The second major error type is image--text mismatch, where the rewritten text is plausible in isolation but does not fully align with the meme's visual template or discourse framing. 
By contrast, pure cultural knowledge gaps are relatively rare. 
This suggests that the main remaining challenge is not simply recognizing culture-specific references, but converting them into humor that is both natural in the target culture and compatible with the accompanying image.
This pattern is consistent across both transcreation directions. 
For ZH$\rightarrow$EN, weak humor reconstruction accounts for 157 of 215 non-top-ranked cases (73.0\%), followed by image--text mismatch (50/215, 23.3\%) and cultural knowledge gaps (8/215, 3.7\%). 
For EN$\rightarrow$ZH, weak humor reconstruction is also the largest category, with 114 of 185 cases (61.6\%), followed by image--text mismatch (58/185, 31.4\%) and cultural knowledge gaps (13/185, 7.0\%).

\subsection{Discussion on Model-Family Scope}

The goal of this work is to validate the proposed framework design rather than to exhaustively benchmark all possible model-family combinations. 
Under this goal, a controlled single-family setting is sufficient to test the contribution of the framework itself, since the main comparison is between different system organizations rather than between different model providers. 
In particular, our strongest baseline B2 already uses a strong single-agent setup, so the comparison directly tests whether explicit role decomposition, cultural planning, and iterative revision provide gains beyond a strong end-to-end alternative. 
This interpretation is further supported by our component analyses, which show that the improvements come from the proposed modules themselves. 
We therefore take the current experiments as evidence for the effectiveness of the framework design under a strongly controlled setting, while broader cross-family validation remains future work.

\section{Related Work}

Prior work on memes mainly falls into three lines: meme understanding, meme generation, and multimodal translation. 

\mypara{Meme Understanding}
Recent advances in vision-language models have strengthened joint reasoning over visual and textual information \cite{he2025artificial}. In social-media settings, multimodal and contextual modeling has been applied to covert advertisement detection, privacy leakage analysis, and misinformation identification \cite{zheng2025chasm,peng2026posts,peng2025infodemic}. More specifically, computational meme research has primarily treated memes as multimodal recognition and interpretation problems, including hate speech and offensiveness detection \cite{kiela2020hateful,suryawanshi2020multioff}, emotion and misogyny analysis \cite{sharma2020semeval,chauhan2020allinone,fersini2022mami}, harmful-content explanation and entity-role reasoning \cite{lin2023unveiling,sharma2023entities}, meme captioning and metaphor interpretation \cite{hwang2023memecap}, intent description \cite{park2024memeintent}, cross-platform contextual grounding \cite{joshi2024contextualizing}, and unified meme understanding with large vision-language models and benchmark corpora \cite{zhao2025memereacon,park2025memeinterpret}. Despite this progress, existing studies largely focus on recognizing, explaining, or moderating meme content rather than adapting it for audiences from different cultural backgrounds \cite{nguyen2024cmu}.

\mypara{Meme Generation}
The second line studies automatic meme generation, from early template-based generation \cite{sadasivam2020memebot} to more recent LLM-based or MLLM-based systems with stronger controllability and interaction \cite{wang2024memecraft,cai2025itermeme}. 
These methods, however, mainly target \emph{in-culture} meme creation rather than cross-cultural re-authoring. 
Multi-agent decomposition and coordination have also been explored in harmful meme detection, structured optimization, and large-scale agent collaboration \cite{liu2025mind,zheng2025gasagent,ye2026evomap}.

\mypara{Multimodal Translation}
Our task is related to multimodal machine translation. Prior work introduces image-grounded benchmarks \cite{elliott2016multi30k}, visual attention \cite{huang2016attention}, imagined visual features \cite{elliott2017imagination}, and image-based pivoting \cite{gella2017pivoting,huang2020pseudo}. However, later studies show that visual context is often underused \cite{wu2021misconceived,li2021sanity,feng2025mnmt}.
Related work also examines text humanization \cite{zheng2025thbench} and detector generalization \cite{liu2025generalization}. In contrast, transcreation emphasizes creative and culturally adapted rewriting for target audiences \cite{diaz2023definition,ho2021marketing,acsova2022advertisements}.

Taken together, these lines do not directly address cross-cultural meme transcreation as a multimodal adaptation problem that must jointly preserve communicative intent, reconstruct humor, and maintain coherence between text and image. 
Most closely related to our work, Zhao et al.~\cite{zhao2026beyond} introduced cross-cultural meme transcreation as a benchmarked multimodal generation task, but their work mainly establishes the task setting without characterizing its core challenges or proposing a tailored method to address them. 
In contrast, we explicitly analyze the core challenges of meme transcreation and develop a tailored framework to address them.

\section{Conclusion}

We study cross-cultural meme transcreation as a multimodal adaptation problem that requires preserving intent, adapting culture-dependent meaning, and maintaining text-image coherence. 
To address this task, we propose a plan-and-revise multi-agent framework with explicit cultural planning, iterative revision, and conditional visual execution. 
Experiments on Chinese-English meme transcreation show that our method consistently outperforms strong baselines, and further analyses confirm the contribution of the main components. 
Overall, our results highlight the importance of explicit cultural reasoning and coordinated multimodal adaptation for effective meme transcreation.

\newpage

\section*{Acknowledgments}

This work was supported by the National Key Research and Development Program of China (No.~2025YFB3110200), the Guangdong Basic and Applied Basic Research Foundation (No.~2026A1515030046), and the State Key Laboratory of Internet Architecture, Tsinghua University (No.~HLW2025ZD14).

% The generated bibliography is included directly so arXiv does not need to
% invoke BibTeX or infer a bibliography workflow.


\begin{thebibliography}{49}

%%% the Web address.

\ifx \showCODEN    \undefined \def \showCODEN     #1{\unskip}     \fi
\ifx \showISBNx    \undefined \def \showISBNx     #1{\unskip}     \fi
\ifx \showISBNxiii \undefined \def \showISBNxiii  #1{\unskip}     \fi
\ifx \showISSN     \undefined \def \showISSN      #1{\unskip}     \fi
\ifx \showLCCN     \undefined \def \showLCCN      #1{\unskip}     \fi
\ifx \shownote     \undefined \def \shownote      #1{#1}          \fi
\ifx \showarticletitle \undefined \def \showarticletitle #1{#1}   \fi
\ifx \showURL      \undefined \def \showURL       {\relax}        \fi
% invisible to TeX
\providecommand\bibfield[2]{#2}
\providecommand\bibinfo[2]{#2}
\providecommand\natexlab[1]{#1}
\providecommand\showeprint[2][]{arXiv:#2}

\bibitem[Shifman(2013a)]%
        {shifman2013memes}
\bibfield{author}{\bibinfo{person}{Limor Shifman}.}
  \bibinfo{year}{2013}\natexlab{a}.
\newblock \showarticletitle{Memes in a Digital World: Reconciling with a
  Conceptual Troublemaker}.
\newblock \bibinfo{journal}{\emph{Journal of Computer-Mediated Communication}}
  \bibinfo{volume}{18}, \bibinfo{number}{3} (\bibinfo{year}{2013}),
  \bibinfo{pages}{362--377}.
\newblock


\bibitem[Shifman(2013b)]%
        {shifman2014memes}
\bibfield{author}{\bibinfo{person}{Limor Shifman}.}
  \bibinfo{year}{2013}\natexlab{b}.
\newblock \bibinfo{booktitle}{\emph{Memes in digital culture}}.
\newblock \bibinfo{publisher}{MIT press}.
\newblock


\bibitem[Ahmed and Masood(2024)]%
        {ahmed2024breaking}
\bibfield{author}{\bibinfo{person}{Saifuddin Ahmed} {and}
  \bibinfo{person}{Muhammad Masood}.} \bibinfo{year}{2024}\natexlab{}.
\newblock \showarticletitle{Breaking Barriers With Memes: How Memes Bridge
  Political Cynicism to Online Political Participation}.
\newblock \bibinfo{journal}{\emph{Social Media + Society}}
  \bibinfo{volume}{10}, \bibinfo{number}{2} (\bibinfo{year}{2024}),
  \bibinfo{pages}{1--12}.
\newblock


\bibitem[Hakok{\"o}ng{\"a}s et~al\mbox{.}(2020)]%
        {hakokongas2020persuasion}
\bibfield{author}{\bibinfo{person}{Eemeli Hakok{\"o}ng{\"a}s},
  \bibinfo{person}{Otto Halmesvaara}, {and} \bibinfo{person}{Inari Sakki}.}
  \bibinfo{year}{2020}\natexlab{}.
\newblock \showarticletitle{Persuasion Through Bitter Humor: Multimodal
  Discourse Analysis of Rhetoric in Internet Memes of Two Far-Right Groups in
  Finland}.
\newblock \bibinfo{journal}{\emph{Social Media + Society}} \bibinfo{volume}{6},
  \bibinfo{number}{2} (\bibinfo{year}{2020}).
\newblock


\bibitem[Bahdanau et~al\mbox{.}(2015)]%
        {bahdanau2015neural}
\bibfield{author}{\bibinfo{person}{Dzmitry Bahdanau},
  \bibinfo{person}{Kyunghyun Cho}, {and} \bibinfo{person}{Yoshua Bengio}.}
  \bibinfo{year}{2015}\natexlab{}.
\newblock \showarticletitle{Neural Machine Translation by Jointly Learning to
  Align and Translate}. In \bibinfo{booktitle}{\emph{Proceedings of the
  International Conference on Learning Representations (ICLR)}}.
\newblock


\bibitem[Specia et~al\mbox{.}(2016)]%
        {specia-etal-2016-shared}
\bibfield{author}{\bibinfo{person}{Lucia Specia}, \bibinfo{person}{Stella
  Frank}, \bibinfo{person}{Khalil Sima'an}, {and} \bibinfo{person}{Desmond
  Elliott}.} \bibinfo{year}{2016}\natexlab{}.
\newblock \showarticletitle{A Shared Task on Multimodal Machine Translation and
  Crosslingual Image Description}. In \bibinfo{booktitle}{\emph{Proceedings of
  the First Conference on Machine Translation: Volume 2, Shared Task Papers}}.
  \bibinfo{publisher}{Association for Computational Linguistics},
  \bibinfo{address}{Berlin, Germany}, \bibinfo{pages}{543--553}.
\newblock


\bibitem[Krishna et~al\mbox{.}(2020)]%
        {krishna-etal-2020-reformulating}
\bibfield{author}{\bibinfo{person}{Kalpesh Krishna}, \bibinfo{person}{John
  Wieting}, {and} \bibinfo{person}{Mohit Iyyer}.}
  \bibinfo{year}{2020}\natexlab{}.
\newblock \showarticletitle{Reformulating Unsupervised Style Transfer as
  Paraphrase Generation}. In \bibinfo{booktitle}{\emph{Proceedings of the 2020
  Conference on Empirical Methods in Natural Language Processing (EMNLP)}}.
  \bibinfo{publisher}{Association for Computational Linguistics},
  \bibinfo{address}{Online}, \bibinfo{pages}{737--762}.
\newblock


\bibitem[Khanuja et~al\mbox{.}(2024)]%
        {khanuja-etal-2024-image}
\bibfield{author}{\bibinfo{person}{Simran Khanuja},
  \bibinfo{person}{Sathyanarayanan Ramamoorthy}, \bibinfo{person}{Yueqi Song},
  {and} \bibinfo{person}{Graham Neubig}.} \bibinfo{year}{2024}\natexlab{}.
\newblock \showarticletitle{An Image Speaks a Thousand Words, but Can Everyone
  Listen? On Image Transcreation for Cultural Relevance}. In
  \bibinfo{booktitle}{\emph{Proceedings of the 2024 Conference on Empirical
  Methods in Natural Language Processing}}. \bibinfo{publisher}{Association for
  Computational Linguistics}, \bibinfo{address}{Miami, Florida, USA},
  \bibinfo{pages}{10258--10279}.
\newblock


\bibitem[Zhao et~al\mbox{.}(2026)]%
        {zhao2026beyond}
\bibfield{author}{\bibinfo{person}{Yuming Zhao}, \bibinfo{person}{Peiyi Zhang},
  {and} \bibinfo{person}{Oana Ignat}.} \bibinfo{year}{2026}\natexlab{}.
\newblock \showarticletitle{Beyond Translation: Cross-Cultural Meme
  Transcreation with Vision-Language Models}.
\newblock \bibinfo{journal}{\emph{arXiv preprint arXiv:2602.02510}}
  (\bibinfo{year}{2026}).
\newblock


\bibitem[Shrout and Fleiss(1979)]%
        {shrout1979intraclass}
\bibfield{author}{\bibinfo{person}{Patrick~E. Shrout} {and}
  \bibinfo{person}{Joseph~L. Fleiss}.} \bibinfo{year}{1979}\natexlab{}.
\newblock \showarticletitle{Intraclass Correlations: Uses in Assessing Rater
  Reliability}.
\newblock \bibinfo{journal}{\emph{Psychological Bulletin}}
  \bibinfo{volume}{86}, \bibinfo{number}{2} (\bibinfo{year}{1979}),
  \bibinfo{pages}{420--428}.
\newblock


\bibitem[Koo and Li(2016)]%
        {koo2016guideline}
\bibfield{author}{\bibinfo{person}{Terry~K. Koo} {and} \bibinfo{person}{Mae~Y.
  Li}.} \bibinfo{year}{2016}\natexlab{}.
\newblock \showarticletitle{A Guideline of Selecting and Reporting Intraclass
  Correlation Coefficients for Reliability Research}.
\newblock \bibinfo{journal}{\emph{Journal of Chiropractic Medicine}}
  \bibinfo{volume}{15}, \bibinfo{number}{2} (\bibinfo{year}{2016}),
  \bibinfo{pages}{155--163}.
\newblock


\bibitem[Kendall and Smith(1939)]%
        {kendall1939problem}
\bibfield{author}{\bibinfo{person}{M.~G. Kendall} {and}
  \bibinfo{person}{B.~Babington Smith}.} \bibinfo{year}{1939}\natexlab{}.
\newblock \showarticletitle{The Problem of $m$ Rankings}.
\newblock \bibinfo{journal}{\emph{The Annals of Mathematical Statistics}}
  \bibinfo{volume}{10}, \bibinfo{number}{3} (\bibinfo{year}{1939}),
  \bibinfo{pages}{275--287}.
\newblock


\bibitem[Cohen(1960)]%
        {cohen1960coefficient}
\bibfield{author}{\bibinfo{person}{Jacob Cohen}.}
  \bibinfo{year}{1960}\natexlab{}.
\newblock \showarticletitle{A Coefficient of Agreement for Nominal Scales}.
\newblock \bibinfo{journal}{\emph{Educational and Psychological Measurement}}
  \bibinfo{volume}{20}, \bibinfo{number}{1} (\bibinfo{year}{1960}),
  \bibinfo{pages}{37--46}.
\newblock


\bibitem[He et~al\mbox{.}(2025)]%
        {he2025artificial}
\bibfield{author}{\bibinfo{person}{Xinlei He}, \bibinfo{person}{Guowen Xu},
  \bibinfo{person}{Xingshuo Han}, \bibinfo{person}{Qian Wang},
  \bibinfo{person}{Lingchen Zhao}, \bibinfo{person}{Chao Shen},
  \bibinfo{person}{Chenhao Lin}, \bibinfo{person}{Zhengyu Zhao},
  \bibinfo{person}{Qian Li}, \bibinfo{person}{Le Yang}, {et~al\mbox{.}}}
  \bibinfo{year}{2025}\natexlab{}.
\newblock \showarticletitle{Artificial intelligence security and privacy: a
  survey}.
\newblock \bibinfo{journal}{\emph{Science China Information Sciences}}
  \bibinfo{volume}{68}, \bibinfo{number}{8} (\bibinfo{year}{2025}),
  \bibinfo{pages}{181101}.
\newblock


\bibitem[Zheng et~al\mbox{.}(2025a)]%
        {zheng2025chasm}
\bibfield{author}{\bibinfo{person}{Jingyi Zheng}, \bibinfo{person}{Tianyi Hu},
  \bibinfo{person}{Yule Liu}, \bibinfo{person}{Zhen Sun},
  \bibinfo{person}{Zongmin Zhang}, \bibinfo{person}{Zifan Peng},
  \bibinfo{person}{Wenhan Dong}, {and} \bibinfo{person}{Xinlei He}.}
  \bibinfo{year}{2025}\natexlab{a}.
\newblock \showarticletitle{{CHASM}: Unveiling Covert Advertisements on Chinese
  Social Media}. In \bibinfo{booktitle}{\emph{Advances in Neural Information
  Processing Systems}}, Vol.~\bibinfo{volume}{38}.
\newblock


\bibitem[Peng et~al\mbox{.}(2026)]%
        {peng2026posts}
\bibfield{author}{\bibinfo{person}{Zifan Peng}, \bibinfo{person}{Yini Huang},
  \bibinfo{person}{Aiwen Lu}, \bibinfo{person}{Qiming Ye},
  \bibinfo{person}{Peixian Zhang}, \bibinfo{person}{Jingyi Zheng},
  \bibinfo{person}{Yule Liu}, \bibinfo{person}{Xuechao Wang},
  \bibinfo{person}{Xinlei He}, {and} \bibinfo{person}{Jiaheng Wei}.}
  \bibinfo{year}{2026}\natexlab{}.
\newblock \showarticletitle{What Your Posts Reveal: A Benchmark and Agentic
  Framework for User-Level Privacy Leakage on Social Media}.
\newblock \bibinfo{journal}{\emph{arXiv preprint arXiv:2606.06784}}
  (\bibinfo{year}{2026}).
\newblock


\bibitem[Peng et~al\mbox{.}(2025)]%
        {peng2025infodemic}
\bibfield{author}{\bibinfo{person}{Zifan Peng}, \bibinfo{person}{Mingchen Li},
  \bibinfo{person}{Yue Wang}, {and} \bibinfo{person}{Daniel~Y. Mo}.}
  \bibinfo{year}{2025}\natexlab{}.
\newblock \showarticletitle{Prompt-Based Contrastive Learning to Combat the
  {COVID-19} Infodemic}.
\newblock \bibinfo{journal}{\emph{Machine Learning}} \bibinfo{volume}{114},
  \bibinfo{number}{1} (\bibinfo{year}{2025}), \bibinfo{pages}{6}.
\newblock


\bibitem[Kiela et~al\mbox{.}(2020)]%
        {kiela2020hateful}
\bibfield{author}{\bibinfo{person}{Douwe Kiela}, \bibinfo{person}{Hamed
  Firooz}, \bibinfo{person}{Aravind Mohan}, \bibinfo{person}{Vedanuj Goswami},
  \bibinfo{person}{Amanpreet Singh}, \bibinfo{person}{Pratik Ringshia}, {and}
  \bibinfo{person}{Davide Testuggine}.} \bibinfo{year}{2020}\natexlab{}.
\newblock \showarticletitle{The Hateful Memes Challenge: Detecting Hate Speech
  in Multimodal Memes}. In \bibinfo{booktitle}{\emph{Advances in Neural
  Information Processing Systems}}, Vol.~\bibinfo{volume}{33}.
  \bibinfo{pages}{2611--2624}.
\newblock


\bibitem[Suryawanshi et~al\mbox{.}(2020)]%
        {suryawanshi2020multioff}
\bibfield{author}{\bibinfo{person}{Shardul Suryawanshi},
  \bibinfo{person}{Bharathi~Raja Chakravarthi}, \bibinfo{person}{Mihael Arcan},
  {and} \bibinfo{person}{Paul Buitelaar}.} \bibinfo{year}{2020}\natexlab{}.
\newblock \showarticletitle{Multimodal Meme Dataset (MultiOFF) for Identifying
  Offensive Content in Image and Text}. In
  \bibinfo{booktitle}{\emph{Proceedings of the Second Workshop on Trolling,
  Aggression and Cyberbullying}}. \bibinfo{pages}{32--41}.
\newblock


\bibitem[Sharma et~al\mbox{.}(2020)]%
        {sharma2020semeval}
\bibfield{author}{\bibinfo{person}{Chhavi Sharma}, \bibinfo{person}{Deepesh
  Bhageria}, \bibinfo{person}{William Scott}, \bibinfo{person}{Srinivas Pykl},
  \bibinfo{person}{Amitava Das}, \bibinfo{person}{Tanmoy Chakraborty},
  \bibinfo{person}{Viswanath Pulabaigari}, {and} \bibinfo{person}{Bj{\"o}rn
  Gamb{\"a}ck}.} \bibinfo{year}{2020}\natexlab{}.
\newblock \showarticletitle{SemEval-2020 task 8: Memotion analysis-the
  visuo-lingual metaphor!}. In \bibinfo{booktitle}{\emph{Proceedings of the
  fourteenth workshop on semantic evaluation}}. \bibinfo{pages}{759--773}.
\newblock


\bibitem[Chauhan et~al\mbox{.}(2020)]%
        {chauhan2020allinone}
\bibfield{author}{\bibinfo{person}{Dushyant~Singh Chauhan}, \bibinfo{person}{SR
  Dhanush}, \bibinfo{person}{Asif Ekbal}, {and} \bibinfo{person}{Pushpak
  Bhattacharyya}.} \bibinfo{year}{2020}\natexlab{}.
\newblock \showarticletitle{All-in-one: A deep attentive multi-task learning
  framework for humour, sarcasm, offensive, motivation, and sentiment on
  memes}. In \bibinfo{booktitle}{\emph{Proceedings of the 1st conference of the
  Asia-Pacific chapter of the association for computational linguistics and the
  10th international joint conference on natural language processing}}.
  \bibinfo{pages}{281--290}.
\newblock


\bibitem[Fersini et~al\mbox{.}(2022)]%
        {fersini2022mami}
\bibfield{author}{\bibinfo{person}{Elisabetta Fersini},
  \bibinfo{person}{Francesca Gasparini}, \bibinfo{person}{Giulia Rizzi},
  \bibinfo{person}{Aurora Saibene}, \bibinfo{person}{Berta Chulvi},
  \bibinfo{person}{Paolo Rosso}, \bibinfo{person}{Alyssa Lees}, {and}
  \bibinfo{person}{Jeffrey Sorensen}.} \bibinfo{year}{2022}\natexlab{}.
\newblock \showarticletitle{SemEval-2022 task 5: Multimedia automatic misogyny
  identification}. In \bibinfo{booktitle}{\emph{Proceedings of the 16th
  International Workshop on Semantic Evaluation (SemEval-2022)}}.
  \bibinfo{pages}{533--549}.
\newblock


\bibitem[Lin et~al\mbox{.}(2023)]%
        {lin2023unveiling}
\bibfield{author}{\bibinfo{person}{Hongzhan Lin}, \bibinfo{person}{Ziyang Luo},
  \bibinfo{person}{Jing Ma}, {and} \bibinfo{person}{Long Chen}.}
  \bibinfo{year}{2023}\natexlab{}.
\newblock \showarticletitle{Beneath the surface: Unveiling harmful memes with
  multimodal reasoning distilled from large language models}. In
  \bibinfo{booktitle}{\emph{Findings of the association for computational
  linguistics: EMNLP 2023}}. \bibinfo{pages}{9114--9128}.
\newblock


\bibitem[Sharma et~al\mbox{.}(2023)]%
        {sharma2023entities}
\bibfield{author}{\bibinfo{person}{Shivam Sharma}, \bibinfo{person}{Atharva
  Kulkarni}, \bibinfo{person}{Tharun Suresh}, \bibinfo{person}{Himanshi
  Mathur}, \bibinfo{person}{Preslav Nakov}, \bibinfo{person}{Md.~Shad Akhtar},
  {and} \bibinfo{person}{Tanmoy Chakraborty}.} \bibinfo{year}{2023}\natexlab{}.
\newblock \showarticletitle{Characterizing the entities in harmful memes: Who
  is the hero, the villain, the victim?}. In
  \bibinfo{booktitle}{\emph{Proceedings of the 17th Conference of the European
  Chapter of the Association for Computational Linguistics}}.
  \bibinfo{pages}{2149--2163}.
\newblock


\bibitem[Hwang and Shwartz(2023)]%
        {hwang2023memecap}
\bibfield{author}{\bibinfo{person}{EunJeong Hwang} {and} \bibinfo{person}{Vered
  Shwartz}.} \bibinfo{year}{2023}\natexlab{}.
\newblock \showarticletitle{MemeCap: A Dataset for Captioning and Interpreting
  Memes}. In \bibinfo{booktitle}{\emph{Proceedings of the 2023 Conference on
  Empirical Methods in Natural Language Processing}}.
  \bibinfo{pages}{1433--1445}.
\newblock


\bibitem[Park et~al\mbox{.}(2024)]%
        {park2024memeintent}
\bibfield{author}{\bibinfo{person}{Jeongsik Park}, \bibinfo{person}{Khoi~P.N.
  Nguyen}, \bibinfo{person}{Terrence Li}, \bibinfo{person}{Suyesh Shrestha},
  \bibinfo{person}{Megan~Kim Vu}, \bibinfo{person}{Jerry~Yining Wang}, {and}
  \bibinfo{person}{Vincent Ng}.} \bibinfo{year}{2024}\natexlab{}.
\newblock \showarticletitle{MemeIntent: Benchmarking intent description
  generation for memes}. In \bibinfo{booktitle}{\emph{Proceedings of the 25th
  Annual Meeting of the Special Interest Group on Discourse and Dialogue}}.
  \bibinfo{pages}{631--643}.
\newblock


\bibitem[Joshi et~al\mbox{.}(2024)]%
        {joshi2024contextualizing}
\bibfield{author}{\bibinfo{person}{Saurav Joshi}, \bibinfo{person}{Filip
  Ilievski}, {and} \bibinfo{person}{Luca Luceri}.}
  \bibinfo{year}{2024}\natexlab{}.
\newblock \showarticletitle{Contextualizing internet memes across social media
  platforms}. In \bibinfo{booktitle}{\emph{Companion Proceedings of the ACM Web
  Conference 2024}}. \bibinfo{pages}{1831--1840}.
\newblock


\bibitem[Zhao et~al\mbox{.}(2025)]%
        {zhao2025memereacon}
\bibfield{author}{\bibinfo{person}{Zhengyi Zhao}, \bibinfo{person}{Shubo
  Zhang}, \bibinfo{person}{Yuxi Zhang}, \bibinfo{person}{Yanxi Zhao},
  \bibinfo{person}{Yifan Zhang}, \bibinfo{person}{Zezhong Wang},
  \bibinfo{person}{Huimin Wang}, \bibinfo{person}{Yutian Zhao},
  \bibinfo{person}{Bin Liang}, \bibinfo{person}{Yefeng Zheng},
  \bibinfo{person}{Binyang Li}, \bibinfo{person}{Kam-Fai Wong}, {and}
  \bibinfo{person}{Xian Wu}.} \bibinfo{year}{2025}\natexlab{}.
\newblock \showarticletitle{MemeReaCon: Probing Contextual Meme Understanding
  in Large Vision-Language Models}. In \bibinfo{booktitle}{\emph{Proceedings of
  the 2025 Conference on Empirical Methods in Natural Language Processing}}.
  \bibinfo{pages}{3559--3582}.
\newblock


\bibitem[Park et~al\mbox{.}(2025)]%
        {park2025memeinterpret}
\bibfield{author}{\bibinfo{person}{Jeongsik Park}, \bibinfo{person}{Khoi P.~N.
  Nguyen}, \bibinfo{person}{Jihyung Park}, \bibinfo{person}{Minseok Kim},
  \bibinfo{person}{Jaeheon Lee}, \bibinfo{person}{Jae~Won Choi},
  \bibinfo{person}{Kalyani Ganta}, \bibinfo{person}{Phalgun~Ashrit Kasu},
  \bibinfo{person}{Rohan Sarakinti}, \bibinfo{person}{Sanjana Vipperla},
  \bibinfo{person}{Sai Sathanapalli}, \bibinfo{person}{Nishan Vaghani}, {and}
  \bibinfo{person}{Vincent Ng}.} \bibinfo{year}{2025}\natexlab{}.
\newblock \showarticletitle{MemeInterpret: Towards an All-in-One Dataset for
  Meme Understanding}. In \bibinfo{booktitle}{\emph{Findings of the Association
  for Computational Linguistics: EMNLP 2025}}. \bibinfo{pages}{16073--16087}.
\newblock


\bibitem[Nguyen and Ng(2024)]%
        {nguyen2024cmu}
\bibfield{author}{\bibinfo{person}{Khoi P.~N. Nguyen} {and}
  \bibinfo{person}{Vincent Ng}.} \bibinfo{year}{2024}\natexlab{}.
\newblock \showarticletitle{Computational Meme Understanding: A Survey}. In
  \bibinfo{booktitle}{\emph{Proceedings of the 2024 Conference on Empirical
  Methods in Natural Language Processing}}. \bibinfo{pages}{21251--21267}.
\newblock


\bibitem[Sadasivam et~al\mbox{.}(2020)]%
        {sadasivam2020memebot}
\bibfield{author}{\bibinfo{person}{Aadhavan Sadasivam}, \bibinfo{person}{Kausic
  Gunasekar}, \bibinfo{person}{Hasan Davulcu}, {and} \bibinfo{person}{Yezhou
  Yang}.} \bibinfo{year}{2020}\natexlab{}.
\newblock \showarticletitle{memeBot: Towards Automatic Image Meme Generation}.
\newblock \bibinfo{journal}{\emph{arXiv preprint arXiv:2004.14571}}
  (\bibinfo{year}{2020}).
\newblock


\bibitem[Wang and Lee(2024)]%
        {wang2024memecraft}
\bibfield{author}{\bibinfo{person}{Han Wang} {and} \bibinfo{person}{Roy Ka-Wei
  Lee}.} \bibinfo{year}{2024}\natexlab{}.
\newblock \showarticletitle{MemeCraft: Contextual and stance-driven multimodal
  meme generation}. In \bibinfo{booktitle}{\emph{Proceedings of the ACM Web
  Conference 2024}}. \bibinfo{pages}{4642--4652}.
\newblock


\bibitem[Cai et~al\mbox{.}(2025)]%
        {cai2025itermeme}
\bibfield{author}{\bibinfo{person}{Yaqi Cai}, \bibinfo{person}{Shancheng Fang},
  \bibinfo{person}{Yadong Qu}, \bibinfo{person}{Xiaorui Wang},
  \bibinfo{person}{Meng Shao}, {and} \bibinfo{person}{Hongtao Xie}.}
  \bibinfo{year}{2025}\natexlab{}.
\newblock \showarticletitle{IterMeme: Expert-Guided Multimodal LLM for
  Interactive Meme Creation with Layout-Aware Generation.}. In
  \bibinfo{booktitle}{\emph{IJCAI}}. \bibinfo{pages}{720--728}.
\newblock


\bibitem[Liu et~al\mbox{.}(2025a)]%
        {liu2025mind}
\bibfield{author}{\bibinfo{person}{Ziyan Liu}, \bibinfo{person}{Chunxiao Fan},
  \bibinfo{person}{Haoran Lou}, \bibinfo{person}{Yuexin Wu}, {and}
  \bibinfo{person}{Kaiwei Deng}.} \bibinfo{year}{2025}\natexlab{a}.
\newblock \showarticletitle{MIND: A multi-agent framework for zero-shot harmful
  meme detection}. In \bibinfo{booktitle}{\emph{Proceedings of the 63rd Annual
  Meeting of the Association for Computational Linguistics (Volume 1: Long
  Papers)}}. \bibinfo{pages}{923--947}.
\newblock


\bibitem[Zheng et~al\mbox{.}(2025b)]%
        {zheng2025gasagent}
\bibfield{author}{\bibinfo{person}{Jingyi Zheng}, \bibinfo{person}{Zifan Peng},
  \bibinfo{person}{Yule Liu}, \bibinfo{person}{Junfeng Wang},
  \bibinfo{person}{Yifan Liao}, \bibinfo{person}{Wenhan Dong}, {and}
  \bibinfo{person}{Xinlei He}.} \bibinfo{year}{2025}\natexlab{b}.
\newblock \showarticletitle{{GasAgent}: A Multi-Agent Framework for Automated
  Gas Optimization in Smart Contracts}.
\newblock \bibinfo{journal}{\emph{arXiv preprint arXiv:2507.15761}}
  (\bibinfo{year}{2025}).
\newblock


\bibitem[Ye et~al\mbox{.}(2026)]%
        {ye2026evomap}
\bibfield{author}{\bibinfo{person}{Qiming Ye}, \bibinfo{person}{Peixian Zhang},
  \bibinfo{person}{Yupeng He}, \bibinfo{person}{Zifan Peng}, {and}
  \bibinfo{person}{Gareth Tyson}.} \bibinfo{year}{2026}\natexlab{}.
\newblock \showarticletitle{Behind {EvoMap}: Characterizing a Self-Evolving
  Agent-to-Agent Collaboration Network}.
\newblock \bibinfo{journal}{\emph{arXiv preprint arXiv:2605.25815}}
  (\bibinfo{year}{2026}).
\newblock


\bibitem[Elliott et~al\mbox{.}(2016)]%
        {elliott2016multi30k}
\bibfield{author}{\bibinfo{person}{Desmond Elliott}, \bibinfo{person}{Stella
  Frank}, \bibinfo{person}{Khalil Sima'an}, {and} \bibinfo{person}{Lucia
  Specia}.} \bibinfo{year}{2016}\natexlab{}.
\newblock \showarticletitle{Multi30K: Multilingual English-German Image
  Descriptions}. In \bibinfo{booktitle}{\emph{Proceedings of the 5th Workshop
  on Vision and Language}}. \bibinfo{pages}{70--74}.
\newblock


\bibitem[Huang et~al\mbox{.}(2016)]%
        {huang2016attention}
\bibfield{author}{\bibinfo{person}{Po-Yao Huang}, \bibinfo{person}{Frederick
  Liu}, \bibinfo{person}{Sz-Rung Shiang}, \bibinfo{person}{Jean Oh}, {and}
  \bibinfo{person}{Chris Dyer}.} \bibinfo{year}{2016}\natexlab{}.
\newblock \showarticletitle{Attention-based Multimodal Neural Machine
  Translation}. In \bibinfo{booktitle}{\emph{Proceedings of the First
  Conference on Machine Translation: Volume 2, Shared Task Papers}}.
  \bibinfo{pages}{639--645}.
\newblock


\bibitem[Elliott and K{\'a}d{\'a}r(2017)]%
        {elliott2017imagination}
\bibfield{author}{\bibinfo{person}{Desmond Elliott} {and}
  \bibinfo{person}{{\'A}kos K{\'a}d{\'a}r}.} \bibinfo{year}{2017}\natexlab{}.
\newblock \showarticletitle{Imagination Improves Multimodal Translation}. In
  \bibinfo{booktitle}{\emph{Proceedings of the Eighth International Joint
  Conference on Natural Language Processing}}. \bibinfo{pages}{130--141}.
\newblock


\bibitem[Gella et~al\mbox{.}(2017)]%
        {gella2017pivoting}
\bibfield{author}{\bibinfo{person}{Spandana Gella}, \bibinfo{person}{Rico
  Sennrich}, \bibinfo{person}{Frank Keller}, {and} \bibinfo{person}{Mirella
  Lapata}.} \bibinfo{year}{2017}\natexlab{}.
\newblock \showarticletitle{Image Pivoting for Learning Multilingual Multimodal
  Representations}. In \bibinfo{booktitle}{\emph{Proceedings of the 2017
  Conference on Empirical Methods in Natural Language Processing}}.
  \bibinfo{pages}{2839--2845}.
\newblock


\bibitem[Huang et~al\mbox{.}(2020)]%
        {huang2020pseudo}
\bibfield{author}{\bibinfo{person}{Po-Yao Huang}, \bibinfo{person}{Junjie Hu},
  \bibinfo{person}{Xiaojun Chang}, {and} \bibinfo{person}{Alexander
  Hauptmann}.} \bibinfo{year}{2020}\natexlab{}.
\newblock \showarticletitle{Unsupervised Multimodal Neural Machine Translation
  with Pseudo Visual Pivoting}. In \bibinfo{booktitle}{\emph{Proceedings of the
  58th Annual Meeting of the Association for Computational Linguistics}}.
  \bibinfo{pages}{8226--8237}.
\newblock


\bibitem[Wu et~al\mbox{.}(2021)]%
        {wu2021misconceived}
\bibfield{author}{\bibinfo{person}{Zhiyong Wu}, \bibinfo{person}{Lingpeng
  Kong}, \bibinfo{person}{Xiang Li}, {and} \bibinfo{person}{Ben Kao}.}
  \bibinfo{year}{2021}\natexlab{}.
\newblock \showarticletitle{Good for misconceived reasons: An empirical
  revisiting on the need for visual context in multimodal machine translation}.
  In \bibinfo{booktitle}{\emph{Proceedings of the 59th Annual Meeting of the
  Association for Computational Linguistics and the 11th International Joint
  Conference on Natural Language Processing (Volume 1: Long Papers)}}.
  \bibinfo{pages}{6153--6166}.
\newblock


\bibitem[Li et~al\mbox{.}(2021)]%
        {li2021sanity}
\bibfield{author}{\bibinfo{person}{Jiaoda Li}, \bibinfo{person}{Duygu Ataman},
  {and} \bibinfo{person}{Rico Sennrich}.} \bibinfo{year}{2021}\natexlab{}.
\newblock \showarticletitle{Vision matters when it should: Sanity checking
  multimodal machine translation models}. In
  \bibinfo{booktitle}{\emph{Proceedings of the 2021 conference on empirical
  methods in natural language processing}}. \bibinfo{pages}{8556--8562}.
\newblock


\bibitem[Feng et~al\mbox{.}(2025)]%
        {feng2025mnmt}
\bibfield{author}{\bibinfo{person}{Yi Feng}, \bibinfo{person}{Chuanyi Li},
  \bibinfo{person}{Jiatong He}, \bibinfo{person}{Zhenyu Hou}, {and}
  \bibinfo{person}{Vincent Ng}.} \bibinfo{year}{2025}\natexlab{}.
\newblock \showarticletitle{Multimodal Neural Machine Translation: A Survey of
  the State of the Art}. In \bibinfo{booktitle}{\emph{Proceedings of the 2025
  Conference on Empirical Methods in Natural Language Processing}}.
  \bibinfo{pages}{22130--22147}.
\newblock


\bibitem[Zheng et~al\mbox{.}(2025c)]%
        {zheng2025thbench}
\bibfield{author}{\bibinfo{person}{Jingyi Zheng}, \bibinfo{person}{Junfeng
  Wang}, \bibinfo{person}{Zhen Sun}, \bibinfo{person}{Wenhan Dong},
  \bibinfo{person}{Yule Liu}, {and} \bibinfo{person}{Xinlei He}.}
  \bibinfo{year}{2025}\natexlab{c}.
\newblock \showarticletitle{{TH-Bench}: Evaluating Evading Attacks via
  Humanizing {AI} Text on Machine-Generated Text Detectors}. In
  \bibinfo{booktitle}{\emph{Proceedings of the 31st ACM SIGKDD Conference on
  Knowledge Discovery and Data Mining V.2}}. \bibinfo{pages}{5948--5959}.
\newblock


\bibitem[Liu et~al\mbox{.}(2025b)]%
        {liu2025generalization}
\bibfield{author}{\bibinfo{person}{Yule Liu}, \bibinfo{person}{Zhiyuan Zhong},
  \bibinfo{person}{Yifan Liao}, \bibinfo{person}{Zhen Sun},
  \bibinfo{person}{Jingyi Zheng}, \bibinfo{person}{Jiaheng Wei},
  \bibinfo{person}{Qingyuan Gong}, \bibinfo{person}{Fenghua Tong},
  \bibinfo{person}{Yang Chen}, \bibinfo{person}{Yang Zhang}, {et~al\mbox{.}}}
  \bibinfo{year}{2025}\natexlab{b}.
\newblock \showarticletitle{On the generalization and adaptation ability of
  machine-generated text detectors in academic writing}. In
  \bibinfo{booktitle}{\emph{Proceedings of the 31st ACM SIGKDD Conference on
  Knowledge Discovery and Data Mining V. 2}}. \bibinfo{pages}{5674--5685}.
\newblock


\bibitem[D{\'i}az-Mill{\'o}n and Olvera-Lobo(2023)]%
        {diaz2023definition}
\bibfield{author}{\bibinfo{person}{Mar D{\'i}az-Mill{\'o}n} {and}
  \bibinfo{person}{Mar{\'i}a~Dolores Olvera-Lobo}.}
  \bibinfo{year}{2023}\natexlab{}.
\newblock \showarticletitle{Towards a Definition of Transcreation: A Systematic
  Literature Review}.
\newblock \bibinfo{journal}{\emph{Perspectives: Studies in Translation Theory
  and Practice}} \bibinfo{volume}{31}, \bibinfo{number}{2}
  (\bibinfo{year}{2023}), \bibinfo{pages}{347--364}.
\newblock


\bibitem[Ho(2021)]%
        {ho2021marketing}
\bibfield{author}{\bibinfo{person}{Nga-Ki~Mavis Ho}.}
  \bibinfo{year}{2021}\natexlab{}.
\newblock \showarticletitle{Transcreation in Marketing: A Corpus-based Study of
  Persuasion in Optional Shifts from English to Chinese}.
\newblock \bibinfo{journal}{\emph{Perspectives: Studies in Translation Theory
  and Practice}} \bibinfo{volume}{29}, \bibinfo{number}{3}
  (\bibinfo{year}{2021}), \bibinfo{pages}{426--438}.
\newblock


\bibitem[{\'A}csov{\'a}(2022)]%
        {acsova2022advertisements}
\bibfield{author}{\bibinfo{person}{Nikola {\'A}csov{\'a}}.}
  \bibinfo{year}{2022}\natexlab{}.
\newblock \showarticletitle{The Transcreation of Advertisements}.
\newblock \bibinfo{journal}{\emph{L10N: Translation and Localization Journal}}
  \bibinfo{volume}{1}, \bibinfo{number}{1} (\bibinfo{year}{2022}),
  \bibinfo{pages}{102--127}.
\newblock


\end{thebibliography}
\end{document}